# RGB-T Tracking via Multi-Modal Mutual Prompt Learning

Yang Luo[1, 2], Xiqing Guo[1, 2], Hui Feng[1], Lei Ao[1]

[1]Aerospace Information Research Institute, Chinese Academy of Sciences, Beijing 100094, China,
[2] University of Chinese Academy of Sciences, Beijing 100040, China.

## Abstract

Object tracking based on the fusion of visible and thermal images, known as RGB-T tracking, has gained increasing attention from researchers in recent years. How to achieve a more comprehensive fusion of information from the two modalities with fewer computational costs has been a problem that researchers have been exploring. Recently, with the rise of prompt learning in computer vision, we can better transfer knowledge from visual large models to downstream tasks. Considering the strong complementarity between visible and thermal modalities, we propose a tracking architecture based on mutual prompt learning between the two modalities. We also design a lightweight prompter that incorporates attention mechanisms in two dimensions to transfer information from one modality to the other with lower computational costs, embedding it into each layer of the backbone. Extensive experiments have demonstrated that our proposed tracking architecture is effective and efficient, achieving state-of-the-art performance while maintaining high running speeds. (The code is available at https://github.com/HusterYoung/MPLT).

## Introduction

With the gradual maturity and popularity of thermal-infrared imaging devices, object tracking based on the fusion of visible and thermal images (RGB-T tracking) has garnered increasing attention from researchers. By incorporating the thermal modality, RGB-T tracking effectively addresses issues related to sensitivity to illumination changes and susceptibility to rain, fog, and other interferences that are commonly encountered in single-modality(visible) tracking. It has found widespread applications in industries such as autonomous driving, intelligent security, and robotics (Xiao et al. 2022). However, researchers have been continuously striving to improve the effectiveness and efficiency of fusing the two modalities to better handle various challenges that may arise during the tracking process, such as illumination variations, fast motion, occlusions, and thermal crossover.

Recently, with the emergence of prompt learning methods in the field of natural language processing, it has become possible to transfer knowledge from pre-trained large models to downstream tasks more quickly and efficiently. Similarly, in the computer vision domain, researchers have also explored prompt learning methods(Jia et al. 2022). Zhu et al.(J. Zhu et al. 2023) applied the concept of prompt learning to RGB-T tracking in their work ViPT. The core idea is to freeze the backbone that is pre-trained on RGB images and integrate the information from the second modality into each layer of the frozen backbone using a set of modality complementarity prompters, as shown in Fig 1(a).

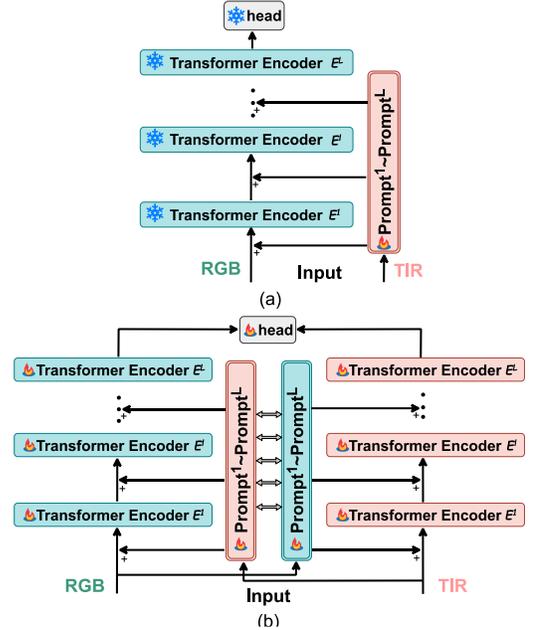

Figure 1 RGB-T tracking framework based on prompt learning, (a) ViPT, (b) Ours(MPLT)

While this method requires training a very small number of parameters, the frozen layers still consume computational resources during training and inference. Therefore, the savings in GPU memory space and computational burden are relatively limited. Additionally, since only unidirectional prompts are used, and the prompter simply calculates weights for pixels within the tokens, this method is actually insufficient for modal fusion.

Therefore, we believe that there exists a better form of modality fusion for RGB-T tracking based on prompt learning. The main challenges lie in balancing the complexity of the modules, training difficulty, and efficiency. Therefore, we have designed a novel approach, as shown in Fig 1(b), based on the idea of mutual prompt learning. In our approach, we introduce two sets of lightweight modules in each layer of the backbone. The fused information is propagated downwards layer by layer through updating strategies, making full and effective use of pre-trained weights to integrate multi-level information from low-level details to high-level semantics across different modalities. Ultimately, this approach enhances the dominant modality information adaptively while suppressing noise from the inferior modality.

To be more specific, we extend the ViT backbone into a dual-branch Siamese architecture. After embedding the images of the two modalities separately, we employ two sets of lightweight prompters, which consist of token attention and spatial attention concatenated together. These modules learn the weights of each token in different modalities and the weights of different spatial positions within each token. Next, we multiply the obtained weights with the original tokens and add them to the corresponding layer's output of the other modality's backbone. Additionally, we add the output of the previous level's prompter to the above, resulting in the prompt output of this layer. We then add this prompt output to the output of the backbone at this layer, which serves as the input for the next layer in the backbone. This process enables multi-level information exchange and integration within the backbone.

Lastly, we explored an online template updating strategy based on classification confidence score and a prediction box correction method based on Kalman filtering to adapt to challenges such as significant appearance variations and severe occlusions during the tracking process. These strategies further enhance the robustness of the tracker.

The evaluation results on multiple publicly available RGB-T datasets demonstrate that our proposed tracker achieves state-of-the-art performance with a relatively small number of trainable parameters.

The main contributions of this article can be summarized as follows:

•Proposed a RGB-T tracking framework, called Multi-Modal Mutual Prompt Learning Tracker (MPLT). By establishing bidirectional modal information interaction channels, it enables the complementary fusion of different modality images during the feature extraction stage, thereby achieving adaptive and precise enhancement of modal information.
•Designed a more efficient Multi-Modal Visual Information Prompter (MVIP). It achieves high-quality information fusion by adaptively generating weights from another modality through the concatenation of two attention mechanisms and adding them to the current modality.
•As a relatively versatile multi-modal fusion tracking architecture, our proposed method can be easily extended to other modal fusion tracking scenarios beyond visible modality.

# Related Works

In this section, we give a brief introduction to RGB-T tracking and visual prompt learning.

## RGB-T Tracking

Currently, there are two main paths followed by RGB-T tracking methods. The first path, based on MDNet(Nam and Han 2016), is adopted by methods such as ((Xiao et al. 2022; Long Li et al. 2019; Lu et al. 2021; 2022; Y. Zhu et al. 2021; Li et al. 2020; Y. Zhu et al. 2019; Gao et al. 2019; Y. Zhu et al. 2022; H. Zhang et al. 2020)). These methods first generate candidate boxes (RoIs) from the search frame, then use specific fusion structures to merge features from different modalities within the RoIs. Finally, they perform binary classification and regress the bbox based on the fused features. Li et al. (Long Li et al. 2019) designed a Multi-Adapter Network that extracts features at three levels: modality-shared features, modality-specific features, and instance-level features. Zhu et al.(Y. Zhu et al. 2022) proposed a three-branch architecture to integrate fused modality features and two modality-specific features, achieving robust target representation. Lu et al. (Lu et al. 2022) proposed a network with a dual-gate structure that utilizes discriminative information from one modality to guide the feature learning of another modality, effectively reducing noise from low-quality modalities. Xiao et al.(Xiao et al. 2022), focusing on the specific challenges of RGBT attributes, employed specific fusion strategies and incorporated three independent Transformer encoders and decoders into each branch to achieve self-enhancement within modalities and interaction across modalities. One important drawback of these methods is that the aspect ratio of the RoI regions is fixed and local. They cannot flexibly adapt to changes in the target's shape and fail to include sufficient background information for feature learning. As a result, the feature interaction between different modality RoIs may be insufficient, leading to inadequate modeling of the global context. This limitation also restricts the mutual enhancement and complementarity between the two modalities(Hui et al. 2023).

The second main path in RGB-T tracking is the Siamese architecture, which is widely acclaimed in visual tracking due to its efficient end-to-end training(Tang, Xu, and Wu 2022). RGB-T tracking based on the Siamese architecture((X. Zhang et al. 2019; 2020; Hui et al. 2023; Luo et al. 2023; Guo et al. 2022; T. Zhang et al. 2021; 2023)) typically involves designing separate feature extraction branches for visible and thermal modalities. Modal fusion modules are introduced in the backbone or after the backbone to fuse the extracted features. The fused features are then fed into the head for classification/regression or directly used for predicting the corners of the target bounding box. These methods generally rely on offline training and do not have an online learning phase. Early Siamese-based RGB-T tracking methods mostly used VGG(Simonyan and Zisserman 2014) or ResNet(He et al. 2016) as the feature extraction backbone. While they achieved high speeds, their accuracy often fell behind MDNet-based methods. In recent years, with the rapid development of the Transformer architecture, more and more RGB-T tracking methods have introduced Transformers as the feature extraction backbone. Hui et al.(Hui et al. 2023) and Luo et al.(Luo et al. 2023) extended the Transformer backbone into the Siamese architecture and achieved modal fusion by inserting multiple cross-attention/self-attention modules into the backbone or placing them after the backbone. Although these methods achieved high accuracy, the stacking of multiple attention modules introduced a large number of parameters, resulting in higher overall complexity of the network.

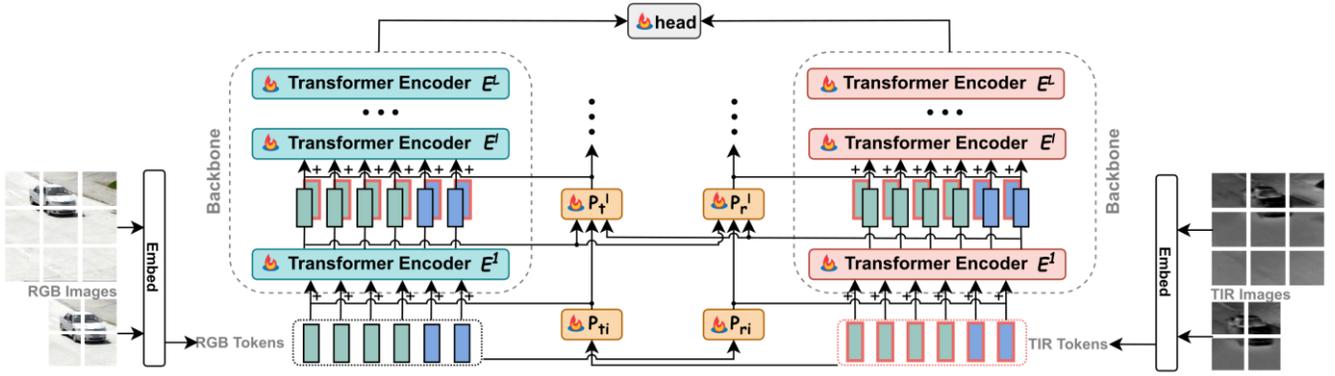

Figure 2 The overall architecture of our proposed RGB-T architecture, with the visible branch on the left and the thermal branch on the right. "+" stands for element-wise addition.

**Visual Prompt Learning**

Prompt, as a form of auxiliary information, has been added to the text to help pre-trained models better adapt to specific downstream tasks. It has been widely applied in the field of NLP(Dai et al. 2023). Recently, researchers have begun to explore the introduction of prompt learning in the computer vision domain. VPT(Jia et al. 2022) was among the first to explore the feasibility of prompt leaning in the visual domain. By freezing the backbone parameters and introducing a small number of learnable parameters in the input space, it achieved comparable downstream performance to full fine-tuning. AIM(Yang et al. 2023) introduced the idea of prompt learning in the domain of video action recognition. By introducing adaptor-based prompters in both spatial and temporal dimensions, it achieved high performance with a small number of trainable parameters. ViPT(J. Zhu et al. 2023) introduced prompt leaning in multi-modal object tracking by integrating limited multi-modal data into a baseline model pre-trained on a large number of RGB images, effectively improving the overall tracking robustness. The aforementioned works based on visual prompt learning all employed the method of freezing the backbone while training the prompt modules. In contrast, in this paper, we adopt the method of Full Finetune + Prompt Leaning. This is because we found in our experiments that freezing the backbone parameters had a limited effect in reducing computational and memory space usage. Training both the backbone and prompt simultaneously allows us to provide additional guidance to the backbone while fine-tuning, which better adapts to the downstream data compared to freezing the backbone.

## Method

The overall architecture of the proposed tracking method is shown in Fig 2, consisting of three components: a dual-branch Transformer backbone, a multi-modal mutual prompting structure, and a localization head. Below, we will provide a detailed description of the workflow of this method.

### Baseline Tracking Model

The basic form of single-object tracking (SOT) involves taking the target region from the initial frame as the template $Z_{RGB} \in \mathbb{R}^{H_z \times W_z \times 3}$, and searching for the template target in the subsequent frame $X_{RGB} \in \mathbb{R}^{H_x \times W_x \times 3}$ to locate the target by bounding it.

For Transformer-based SOT models like the baseline(Ye et al. 2022)model, the first step is to convert $Z_{RGB}, X_{RGB}$ into patches of size $P \times P$ through embedding:

$$\{Z_{RGB}, X_{RGB}\} \rightarrow \{Z_{RGB}^P \in \mathbb{R}^{N_z \times D}, X_{RGB}^P \in \mathbb{R}^{N_x \times D}\} \quad (1)$$

Where $N_z = H_z W_z / P^2$, $N_X = H_X W_X / P^2$, $D = P^2 \times C$ ($C$ is the number of image channels, here is 3). Next, $Z_P$, $X_P$ are concatenated and fed into the backbone to learn features and facilitate interaction between the template and search regions:

$$H_{RGB} = Concat(Z_{RGB}, X_{RGB}) \quad (2)$$

$$B_{RGB} = h\big(f(H_{RGB})\big) \quad (3)$$

Where $H_{RGB} \in \mathbb{R}^{N \times D}, N = N_X + N_Z, f$ represent the ViT backbone, $h$ represent the localization head, $B_{RGB}$ represent the final output bounding box.

## Mutual Prompt Learning for RGB-T Tracking

**Overview** In our approach, in addition to the visible modality, we also introduce the thermal modality. Therefore, we have:

$$\{Z_{TIR}, X_{TIR}\} \rightarrow \{Z_{TIR}^P \in \mathbb{R}^{N_z \times D}, X_{TIR}^P \in \mathbb{R}^{N_x \times D}\} \quad (4)$$

$$H_{TIR} = Concat(Z_{TIR}, X_{TIR}) \quad (5)$$

Moreover, we extend the backbone into a dual-branch Siamese architecture, where $f_{RGB}$ represents the backbone for extracting RGB features, and $f_{TIR}$ represents the backbone for extracting thermal features. Let $E_{RGB}^l$ denote the encoder in the l-th layer of $f_{RGB}$, and $E_{TIR}^l$ denote the encoder in the l-th layer of $f_{TIR}$ (with a total of L layers). $H_{RGB}^l$ represents the output of $E_{RGB}^l$, and $H_{TIR}^l$ represents the output of $E_{TIR}^l$:

$$H_{RGB}^l = E^l(H_{RGB}^{l-1}), l = 1,2,...,L \quad (6)$$

Before the input to the first layer of the encoder, the tokens from the visible modality and the thermal modality (denoted as $H_{RGB}^{init}$ and $H_{TIR}^{init}$, respectively) are first passed through an Initial Multi-Modal Visual Information Prompter (IMVIP). The outputs of the IMVIP are then added to the tokens of the other modality:

$$H_{RGB}^0 = H_{RGB}^{init} + P_{ti}$$
$$P_{ti} = IMVIP(H_{RGB}^{init}, H_{TIR}^{init})$$
$$H_{TIR}^0 = H_{TIR}^{init} + P_{ri}$$
$$P_{ri} = IMVIP(H_{TIR}^{init}, H_{RGB}^{init}) \quad (7)$$

Where $P_{ti}$ and $P_{ri}$ represent the output of the IMVIP. In the subsequent layers of the encoder, we have:

$$H_{RGB}^l = H_{RGB}^{l-1} + P_t^l$$
$$P_t^l = MVIP(H_{RGB}^{l-1}, P_t^{l-1}, H_{TIR}^{l-1})$$
$$H_{TIR}^l = H_{TIR}^{l-1} + P_r^l$$
$$P_r^l = MVIP(H_{TIR}^{l-1}, P_r^{l-1}, H_{RGB}^{l-1}) \quad (8)$$

Where $l = 1,2,...,L$, $MVIP$ represents the Multi-Modal Visual Information Prompter, and $P_t^l$ and $P_r^l$ represent the output of the MVIP in the l-th layer(with $P_t^0 = P_{ti}$ and $P_r^0 = P_{ri}$).

Lastly, the features from the two branches, after mutual prompt learning, are concatenated. Then, they are passed through a linear layer to reduce the channel dimensionality before being fed into the localization head for classification prediction and target box regression:

$$B = h\left(DR\left(Concat(H_{RGB}^L, H_{RGB}^L)\right)\right) \quad (9)$$

Where $DR$ represents the linear layer for dimensionality reduction. The details of the localization head $h$ can be referred to OSTrack(Ye et al. 2022).

**Multi-Modal Visual Information Prompter** In VPT(Jia et al. 2022), several forms of visual prompts were explored, and the authors empirically demonstrated that converting the prompts into token form and inserting it into the original token sequence yielded the best performance in downstream tasks. It is well known that the parameter count of the Transformer structure is related to the number of tokens, so this form of prompt insertion comes with a higher computational burden. In our task, we use two modalities of images with the same size for prompt learning. Therefore, considering efficiency, we adopt the second visual prompt method mentioned in(Jia et al. 2022) , which directly superimposes the prompt information on the original tokens. This approach does not increase the token count while achieving comparable performance in downstream tasks to the first method.

In terms of the specific structure of the prompter, the proposed modules can be divided into two subclasses. The first one is the Initial Multi-Modal Visual Information Prompter (IMVIP), which has only two input branches (as shown in Equation 7). The second one is the Multi-Modal Visual Information Prompter (MVIP), which is designed for intermediate layers and has three input branches (as shown in Equation 8). Since the basic structure of the two types of prompter is the same except for the number of branches, we will now focus on providing a detailed introduction to the MVIP module.

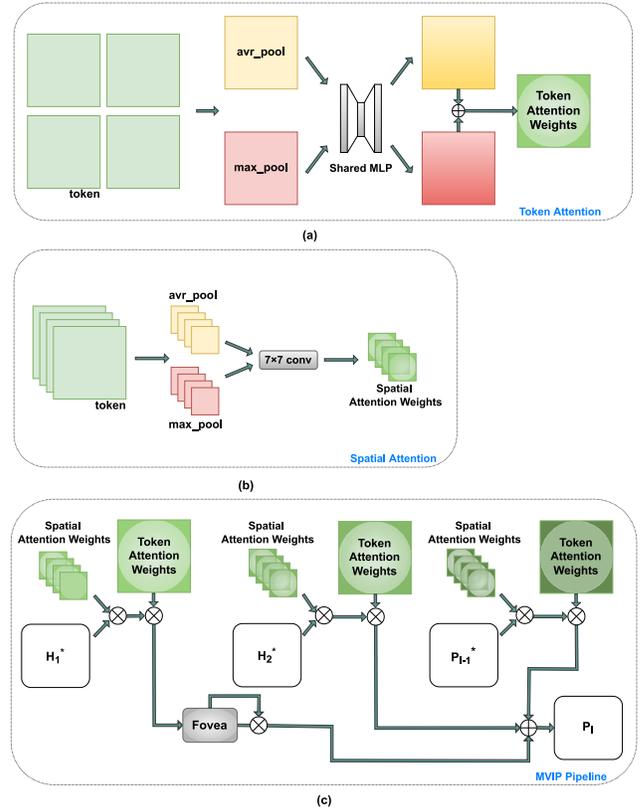

Figure 3 The overall architecture of Multi-Modal Visual Information Prompter (MVIP), (a) is the process of generating token attention weights, (b) is the process of generating spatial attention weights, (c) is the overall pipeline.

As shown in Fig 3, the MVIP module has three input branches: the output of the encoder of current modality from the previous layer, the output of the encoder of the other modality from the previous layer, and the output of the previous MVIP. Inspired by CBAM(Woo et al. 2018), the workflow of the MVIP module consists of three steps:

(1) Spatial attention operations are performed on each of the three input branches. Specifically, for a particular branch, let's assume the input tokens are represented by $H$, average pooling and max pooling are first applied along the $D$ dimension:

$$W_{avr}^s = avr\_pool(H, dim \sim D) \quad (10)$$

$$W_{max}^s = max\_pool(H, dim \sim D) \quad (11)$$

Next, $W_{avr}^s$ and $W_{max}^s$ undergo channel dimension reduction in the $N$ dimension using a 1×1 convolutional layer $g_{s1}$. This projects the features into a lower-dimensional latent embedding. The resulting features are then passed through a ReLU layer for non-linear enhancement before being projected back to the original dimension using another 1×1 convolutional layer $g_{s2}$. This generates a weight map with the same size as the token dimension $D$:

$$W_{spatial} = g_{s2}\left(relu(g_{s1}(W_{avr}^s))\right) + g_{s2}\left(relu(g_{s1}(W_{max}^s))\right) \quad (12)$$

Finally, $W_{spatial}$ is multiplied element-wise with the original tokens to obtain the tokens with redistributed weights in the spatial dimension:

$$H^* = H * W_{spatial} \quad (13)$$

(2) Token attention operations are performed on each of the three input branches. Specifically, for a particular branch, the average and maximum values are computed along the $N$ dimension of the tokens:

$$W_{avr}^t = mean(H, dim \sim N) \quad (14)$$

$$W_{max}^t = max(H, dim \sim N) \quad (15)$$

Then, $W_{avr}^t \in \mathbb{R}^{1 \times D}$ and $W_{max}^t \in \mathbb{R}^{1 \times D}$ are concatenated along the first dimension, and a 7×7 convolutional layer $g_t$ with padding is used to reduce the first dimension back to 1, resulting in a $D$-dimensional weight map $W_{token}$:

$$W_{token} = g_t \cdot Concat(W_{avr}, W_{max}) \quad (16)$$

Then, the weight map is multiplied element-wise with the input tokens to obtain the token sequence with reweighted values:

$$H^* = H * W_{token} \quad (17)$$

(3) Execute the spatial fovea operation on the token of current modality, which first applies a λ-smoothed spatial softmax across all the spatial dimensions, and produces the enhanced embeddings $H_{RGB}^*$ by applying the channel-wise spatial attention-like mask $W_{fovea}$ over $H_{RGB}^*$. Then, the tokens from the other two branches are added to obtain the output of MVIP:

$$P_l = W_{fovea} \cdot H_1^* + H_2^* + P_{l-1}^* \quad (18)$$

Where $H_1^*$, $H_2^*$ represent the outputs of the tokens from two different modalities after the two attention operations, and $P_{l-1}^*$ represents the output of the previous MVIP after the two attention operations.

**Discussion** Regardless of token attention or spatial attention, their purpose is to adaptively allocate weights to different tokens or pixel positions within tokens, assuming that the image information is redundant. This aims to enhance the target information and suppress noise. In previous works on multi-modal image fusion using visual prompts, such as ViPT(J. Zhu et al. 2023), the prompters were designed considering only the pixel positions within tokens, and the prompter structure was relatively simple, which failed to accurately extract the modal information for prompting. This led to poor performance of the model in challenging sequences that require effective utilization of complementary thermal modal information, such as low-illumination conditions (as shown in Table 2). Additionally, considering the complementary nature of information between visible and thermal modalities, there are cases where the visible modality needs supplementation from the thermal modality, and vice versa. Therefore, the design of visual prompts should not be unidirectional but rather bidirectional. A mechanism based on mutual prompt learning for modality information transfer can more fully capture various complementary situations.

---

**Algorithm 1**: Online Template Update and Kalman Filter Based Prediction Correction

**Input**: The predicted bbox and classification confidence of the last n frames.
**Parameter**: n, thr$^u$, thr$^b$
**Output**: Bbox to be corrected.
    **while** Current frame number<Total frame number **do**
        Record the predicted bbox and classification confidence of the last n frames
        **if** Current confidence>thr$^u$ **then**
            Re-crop the template
        **else** Do nothing
        **if** Current confidence<thr$^b$ **then**
            According to the prediction of frame 0~n-1, use the Kalman filter to predict the bbox of the n$^{th}$ frame, and use it to correct the network prediction results.
        **else** Do nothing
    **end while**

## Inference

Object tracking, as a real-time task, also considers temporal information. Therefore, in the inference stage, to incorporate temporal information, we designed a simple online template update method based on the confidence scores from the classification head. Additionally, we combined it with Kalman filtering to correct low-confidence prediction results. The specific algorithm is presented in Algorithm 1.

# Experiments

This section mainly introduces the specific implementation method of MPLT, evaluates its performance on multiple datasets, and conducts ablation experiments on the involved modules/methods to verify its effectiveness.

**Implementation Details**
Our model is implemented using PyTorch (Paszke et al. 2019) and trained using two RTX 3090 GPUs. The batchsize is set to 24, and each epoch samples 60,000 images. We train the model for a total of 15 epochs on LasHeR training set. The learning rate for the backbone is set to 7.5e-5, while the learning rate for other parts is set to 7.5e-4. After the 10th epoch, the learning rate is decayed by a factor of 10. We use the AdmW(Loshchilov and Hutter 2017) optimizer for iteration with a weight decay of 1e-4. The input image sizes for the network are as follows: search frame is 256×256, and template frame is 128×128. The structure and parameter settings of the loss function are the same as OSTrack (Ye et al. 2022). The template update and prediction correction consider 16 frames, and the confidence thresholds are set to $thr^u$=0.91 and $thr^b$=0.25. Additionally, for evaluation metrics, we use commonly used precision/success rate(PR/SR) metrics and set the center location error (CLE) threshold to the conventional value of 20 pixels.

Table 1. Overall performance on RGBT234 dataset.

| Trackers | PR | SR |
|---|---|---|
| MPLT | 88.4% | 65.7% |
| MACFT(Luo et al. 2023) | 85.7% | 62.2% |
| OSTrack(Ye et al. 2022) | 72.9% | 54.9% |
| ViPT(J. Zhu et al. 2023) | 83.5% | 61.7% |
| DMCNet(Lu et al. 2022) | 83.9% | 59.3% |
| TBSI(Hui et al. 2023) | 87.1% | 63.7% |
| APFNet(Xiao et al. 2022) | 82.7% | 57.9% |

**Evaluation on RGBT234 Dataset**
The RGBT234 dataset(Li et al. 2019) consists of 234 sequences with approximately 116.7K frames. From Table 1, it can be observed that our proposed algorithm achieves the best tracking performance among all state-of-the-art RGB-T tracking algorithms. It outperforms the second and third-ranked algorithms by 1.3%/2% & 2.7%/3.5% in terms of PR and SR respectively. Compared to the baseline model, our model achieved an improvement of 15.5% in terms of PR and 10.8% in terms of SR. These results provide strong evidence for the effectiveness of our algorithm.

**Evaluation on LasHeR Dataset**
The LasHeR dataset(Li et al. 2022) is currently the largest RGB-T tracking dataset with precise annotation and alignment. The dataset is divided into training and testing sets, and it presents a higher level of tracking difficulty compared to the RGBT234 dataset. We evaluate the trackers on 245 test video sequences in terms of precision plot and success plot. The results are reported in Fig 4. Our tracker achieves state-of-the-art performance in terms of PR and SR. It outperforms the second and third-ranked trackers by 1.5%/0.8% and 6.7%/4.7% respectively. Compared to the baseline model, our model achieved an improvement of 22.2% in terms of PR and 17.6% in terms of SR. These results further validate the effectiveness of our algorithm.

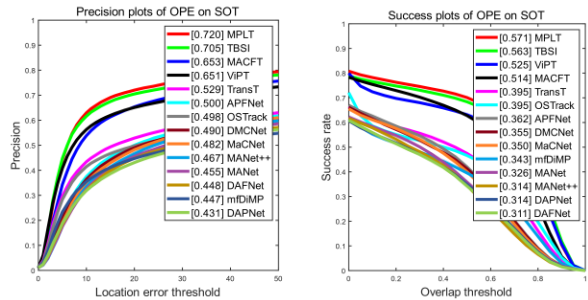

Figure 4 Overall performance on LasHeR test set.

**Attribute-Based Evaluation** In order to better evaluate the complementary fusion capability of our proposed tracking approach across two modalities, we selected 5 challenging attributes from the LasHeR dataset, namely low illumination(LI), high illumination(HI), thermal crossover(TC), frame loss(FL), and fast motion(FM). Comparisons with the second to the fourth-ranked trackers in terms of SR metrics revealed that our proposed model achieved the best performance in these challenges, providing ample evidence of its ability to fuse complementary information between the two modalities, and proving its high robustness even in scenarios with target blurring or loss, too.

Table 2. Attribute-based evaluation on LasHeR dataset.

| Attributes | MPLT | ViPT | TBSI | MACFT |
|---|---|---|---|---|
| TC | 50.5% | 46.0% | 50.1% | 44.4% |
| LI | 49.6% | 41.2% | 49.3% | 45.1% |
| HI | 59.7% | 54.2% | 58.2% | 59.4% |
| FL | 49.4% | 46.9% | 47.5% | 46.2% |
| FM | 56.6% | 51.5% | 55.7% | 50.8% |

**Evaluation on RGBT210 Dataset**
The RGBT210(Li et al. 2017) dataset consists of 210 sequences with approximately 104.8K frames. As shown in Table 3, our tracker achieved the best performance in terms of PR/SR metrics for RGBT210 dataset.

Table 3. Overall performance on RGBT210 dataset.

| Trackers | PR | SR |
|---|---|---|
| MPLT | 86.2% | 63.0% |
| CAT(Li et al. 2020) | 79.2% | 53.3% |
| DMCNet(Lu et al. 2022) | 79.7% | 55.5% |
| mfDiMP(L. Zhang et al. 2019) | 84.9% | 59.3% |
| TBSI(Hui et al. 2023) | 85.3% | 62.5% |

**Ablation Study**
In this section, each component of MPLT will be analyzed separately to validate their effectiveness.

**Variants Comparison** We further explored different variant versions of MPLT. (1)MPLT-Full: Full MPLT model. (2) MPLT-w/o MVIP: Remove all MVIP modules, Concatenate the features from the two backbones directly, and feed them into the localization head. (3) MPLT-w/o SA: Remove all spatial attention operations in MVIP. (4) MPLT-w/o TA: Remove all token attention operations in MVIP. (5) MPLT-w/o TU: Remove template update operation. (6) MPLT-w/o KF: Remove the prediction refinement step based on Kalman filtering. As shown in Table 4, It can be observed that the modules/methods we designed have improved the tracking performance to varying degrees.

Table 4. Ablation studies on LasHeR test set.

| Variants | PR | SR |
|---|---|---|
| MPLT-Full | 72.0% | 57.1% |
| Baseline(OSTrack) | -22.2% | -17.6% |
| MPLT-w/o MVIP | -6.1% | -5.6% |
| MPLT-w/o SA | -4.7% | -3.6% |
| MPLT-w/o TA | -4.6% | -3.5% |
| MPLT-w/o TU | -0.4% | -0.3% |
| MPLT-w/o KF | -0.2% | -0.2% |

**Freeze or Unfreeze?** To investigate the impact of freezing the backbone on our method, we introduced a control group called MPLT-F. The network architecture of MPLT-F is the same as MPLT, but the parameters of the backbone will be frozen. Similarly, ViPT-UF in the control group represents unfreezing the backbone parameters of ViPT.

As shown in Table 5, freezing the backbone parameters significantly reduces the performance of our model, but it still outperforms ViPT, which is based on unidirectional prompt learning. This demonstrates the effectiveness of the prompt structure we designed. Additionally, it is worth noting that freezing the backbone does not significantly reduce the GPU memory utilization compared to trainable parameters. This implies that full finetuning remains a cost-effective choice. Lastly, compared to TBSI, which uses stacked cross-attention modules to fuse tokens, our model is more efficient in terms of parameter count and FLOPS.

Table 5. Comparison of several ViT-B-based RGB-T trackers with and without freezing backbone parameters. Params[†] denotes the number of trainable parameters. GPU Usage[‡] refers to the GPU memory usage during model training under the condition of using the same batchsize (set MPLT to 1, and other models are scaled proportionally).

| Index | MPLT | MPLT-F | ViPT | ViPT-UF | TBSI |
|---|---|---|---|---|---|
| GPU Usage[‡] | 100% | 97% | 45% | 55% | 96% |
| Params[†] | 97M | 11.9M | 0.84M | 87.6M | 307M |
| FLOPS | 58.7G | 58.7G | 21.3G | 21.3G | 82.2G |
| PR(LasHeR) | 72.0% | 68.2% | 65.1% | 64.8% | 70.5% |
| SR(LasHeR) | 57.1% | 54.0% | 52.5% | 51.7% | 56.3% |
| FPS(LasHeR) | 22.8 | - | 38.5 | - | 36.2 |

**Visualization**
To demonstrate the robustness of our model in challenging tracking scenarios, we visualized the attention map between the template and the search region, as shown in Fig 5. Compare to the baseline model, our model accurately matches the corresponding regions between the template and the search image in high/low-illumination conditions and thermal-crossover scenarios, etc. Additionally, it effectively suppresses noise from low-quality modalities.

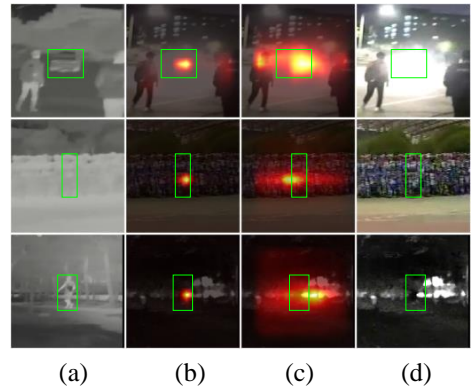

(a)　　　(b)　　　(c)　　　(d)

Figure 5 Visualization of attention maps. (a) TIR search region. (b) Attention map-MPLT. (c) Attention map-Baseline. (d) RGB search region.

## Conclusion

In this work, we propose MPLT, an RGB-T tracking method based on multi-modal mutual prompt learning. MPLT effectively extracts and fuses complementary information from different modalities while maintaining a low computational cost. It successfully transfers the object tracking foundation model trained on single-modal images to downstream tasks, achieving high performance. Extensive experiments provide ample evidence that our method is both effective and efficient.

## Acknowledgements

This work was supported by The National Key Research and Development Program of China (No.2022YFB3901800, No.2022YFB3901805).